\documentclass[runningheads]{llncs}

\usepackage{authblk,times,latexsym,amsmath,graphicx,comment,url,todonotes,multicol,arabtex,utf8,latexsym}
\usepackage{subcaption}
\begin{document}
\title{Code-switching Language Modeling With Bilingual Word Embeddings: \\ A Case Study for Egyptian Arabic-English}
\titlerunning{Code-switching Language Modeling With Bilingual Word Embeddings}
\author{Injy Hamed\inst{1}\and
Moritz Zhu\inst{2}\and
Mohamed Elmahdy\inst{3}\and
Slim Abdennadher\inst{1}\and\newline
Ngoc Thang Vu\inst{2}}

\authorrunning{I. Hamed et al.}
\institute{Computer Science Department, The German University in Cairo, Cairo, Egypt \\
\email{\{injy.hamed,slim.abdennadher\}@guc.edu.eg}\and
Institute for Natural Language Processing, University of Stuttgart, Stuttgart, Germany\\
\email{\{moritz.zhu,thang.vu\}@ims.uni-stuttgart.de}\and
Data Science Department, Raisa Energy LLC, Cairo, Egypt\\
\email{melmahdy@raisaenergy.com}
}
\maketitle              % typeset the header of the contribution
\begin{abstract}
Code-switching (CS) is a widespread phenomenon among bilingual and multilingual societies. The lack of CS resources hinders the performance of many NLP tasks. In this work, we explore the potential use of bilingual word embeddings for code-switching (CS) language modeling (LM) in the low resource Egyptian Arabic-English language. We evaluate different state-of-the-art bilingual word embeddings approaches that require cross-lingual resources at different levels and propose an innovative but simple approach that jointly learns bilingual word representations without the use of any parallel data, relying only on monolingual and a small amount of CS data. While all representations improve CS LM, ours performs the best and improves perplexity 33.5\% relative over the baseline. 
\end{abstract}

\section{Introduction}
\label{sec:intro}
Code-switching is a common phenomenon in multilingual communities where people use more than one language in a conversation \cite{poplack1978syntactic}. Due to several factors such as colonization, the rise in education levels and international business and communication, code-switching is seen in several Arab countries, such as Arabic-French in Morocco, Tunisia, Algeria, and Lebanon and Arabic-English in Egypt, Jordon and Saudi Arabia. CS is becoming widely used in Egypt, especially among urban youth, which has motivated research in the NLP field in that direction \cite{HEA17,SEA19}. As shown in~\cite{HEA18}, Egyptians mix the three languages: Modern Standard Arabic, Dialectal Arabic and English, thus posing challenges to NLP tasks. With the widespread of CS due to globalization, more attention from the speech and language research community has been given towards building NLP applications that can handle such mixed-language input. However, given the scarcity of CS data, NLP tools often fail or need to be extensively adapted to perform well on CS data \cite{CSV16}, including the LM task. Language modeling (LM) is a widely-used technique in many NLP applications, including Automatic Speech Recognition (ASR) systems. The performance of language models on code-switched data are greatly hindered by data sparsity problem. %the scarce training data. 
The problem of data sparseness affects the performance of traditional n-grams and neural-based LMs, as many word sequences can occur in the testing data without being present in the training data. \\

Previous work proposed several techniques such as artificial CS text generation using statistical machine translation-based methods~\cite{VLW+12} and integrating linguistic knowledge in recurrent neural networks and factored LMs~\cite{AVS13} or to pose constraints on CS boundaries~\cite{YF14b}. Another option is leveraging multi-task learning, where the model jointly predicts the next word and POS tagging on CS text. Recently,~\cite{GPJ18} proposed dual LMs, where two complementary monolingual LMs are trained separately and then a probabilistic model is used to switch between them. This approach overcame the problem of limited CS data by relying on the large amounts of monolingual data.

In this paper, we address the data sparseness problem in CS LM from a different perspective, leveraging the advantages of representing words using continuous vectors~\cite{RHW86,BDV+03,MCC13}. We explore the use of bilingual word embeddings \footnote{The bilingual word embeddings and the compiled Egyptian Arabic-English dictionary and thesaurus can be obtained by contacting the authors.} as shared latent space to bridge the gap between languages in CS LM for Egyptian Arabic-English. Compared to other previous work, our proposed method does not require any external knowledge, e.g. generated from a part-of-speech tagger or a syntax parser which is not a trivial task in the CS context \cite{CSV16}. 
To the best of our knowledge, this work is the first research towards this direction.
We compare different bilingual embeddings using state-of-the-art approaches that rely on different levels of cross-lingual supervision; word-aligned~\cite{LPM15} and sentence-aligned~\cite{HB14} parallel corpora, and a bilingual lexicon~\cite{FD14}. Moreover, we propose two new approaches, where the first approach only relies on monolingual and small CS corpora (Bi-CS) and the second approach combines two of the existing approaches \cite{FD14,LPM15}. We investigate their impact on LM as well as evaluate them intrinsically on monolingual and bilingual tasks. 
Our results reveal that bilingual embeddings improve LM, with our proposed bilingual embeddings (Bi-CS) performing best, achieving 33.5\% relative improvement in perplexity (PPL) over the baseline.

\section{Related Work}
\label{sec:relatedWork}
Bilingual word embeddings have proven to be a valuable resource to various NLP tasks, such as machine translation \cite{WZP+18}, cross-lingual entity linking \cite{TS16}, document classification \cite{GBC15}, cross-lingual information retrieval \cite{VM15}, part-of-speech tagging \cite{GS15} and sentiment analysis \cite{WKP16}. Several approaches have been proposed for building bilingual word embeddings, where the bilingual word representations across multiple languages can be jointly learned, or where independently-learned monolingual representations can be mapped to one vector space. For both tasks, different forms of cross-lingual supervision are leveraged, including alignments at word level \cite{HB14}, sentence level \cite{HB14,GBC15}, both word and sentence level \cite{LPM15}, or document level \cite{VM15,VM16}, in addition to bilingual lexicons \cite{VM13,FD14,GS15,WKP16} and comparable un-aligned data \cite{VM16}. A comprehensive survey on crosslingual word embedding models is provided by Ruder et al. \cite{RVVS17}. The survey presents a comparison between the models regarding their data requirements and objective functions, as well as a discussion covering the different evaluation methods used for cross-lingual word embeddings.\\

In \cite{UFD+16}, Upadhyay et al. present an extensive evaluation of four popular cross-lingual embedding methods \cite{LPM15,HB14,FD14,VM15a} that all require parallel training data, but differ in the degree of data parallelism required. In this work, we propose a new approach (Bi-CS) for training bilingual word embeddings that requires no level of cross-lingual supervision and compare it against the first three models compared in \cite{UFD+16} on two tasks: language modeling and concept categorization. While the existing approaches require different levels of cross-lingual supervision (word-aligned \cite{LPM15} and sentence-aligned \cite{HB14} parallel corpora as well as bilingual lexicon \cite{FD14}), Bi-CS only uses monolingual data in addition to a small amount of CS data. We also investigate integrating two of the existing approaches \cite{FD14,LPM15}.

\section{Data}
\paragraph{CS Data} For language modeling evaluation, we further extend the Egyptian Arabic speech transcriptions obtained in \cite{HEA18}.
The corpus contains a total of 14,191 Arabic and 7,758 English words, which shows high usage of the embedded English language in the conversations. Out of the total 2,407 sentences, there are 573 (23.8\%) monolingual Arabic, 239 (9.9\%) monolingual English and 1,595 (66.3\%) CS sentences, which also shows a high rate of code-mixing. A sample of the the corpus is given in Table  \ref{table:samples}.
\setcode{utf8}
\begin{table}
    \centering
    \begin{tabular}{|p{12cm}|}
    \hline
    \multicolumn{1}{|c|}{\textbf{Sentences}} \\ \hline
              $\ast$\textit{ Actually}
          \< هتبقى>
        \textit{experience}
        \< حلوة ان انت ت  >
         \textit{do research }
        \< .في مصر >\\
           Translation: \textit{Actually} it will be a nice \textit{experience} that you \textit{do research} in Egypt.            \\ \hline
           \vspace{-0.2cm}\RL{$\ast$ \< لو انت عندك>
        \LR{\textit{object that you need to track in a fish-eye camera}}
        \<بتبقى اصعب علشان ال >
         \LR{\textit{distortion}}
        \<اللي في الجناب>
         \LR{\textit{ok}}
        ؟}\\
          Translation:  If you have an \textit{object that you need to track in a fish-eye camera} it becomes harder due to the \textit{distortion} in the edges \textit{ok}?           \\ \hline
        \vspace{-0.2cm}\RL{$\ast$ \< كانت عن ان احنا بنحاول ن  >
        \LR{\textit{convert visual information into sounds for visually impaired people.}}}\\
            Translation:   It was about that we were trying to \textit{convert visual information into sounds for visually impaired people.}          \\ \hline
    \end{tabular}
    \caption{Samples from the Egyptian Arabic-English CS speech corpus. The $\ast$ marks the start of the sentence.}
    \label{table:samples}
\end{table}

\paragraph{Resources for Bilingual Embeddings} For word embeddings training, we gathered text from Facebook pages that are related to Egypt and tweets obtained from Twitter with the location restricted to Cairo. The corpus contains a total of 1,521,818 monolingual Arabic, 270,741 monolingual English and 123,445 CS sentences, as further detailed in Table \ref{table:sm_corpus}. We obtain parallel sentences from LDC's BBN Arabic-Dialect/English Parallel Text~\cite{ZMD+12}, containing 38,154 Egyptian Arabic-English aligned sentences. All the embeddings were trained using the text corpus and the parallel corpus. For BiCCA, we obtain 41,777 Egyptian Arabic-English translation pairs from \textit{Lisaan Masy}\footnote{\url{http://eg.lisaanmasry.com/intro_en/index.html}} dictionary, out of which 14,812 translation pairs were found in our text corpus. We extracted the text from the available PDF format and parsed it into a machine-readable format. We also extracted an Egyptian Arabic-English thesaurus provided in the \textit{Lisaan Masry} dictionary for the intrinsic evaluation. The extracted thesaurus contains a total of 43 general categories, divided into 356 sub-categories, each having an average of 35 Arabic and 29 English words. After pruning to words available in the text corpus, we end up with 40 general categories, 343 sub-categories, with an average of 25 Arabic and 10 English words in each sub-category.

\begin{table}[]
    \centering
\begin{tabular}{|l|r|r|r|r|}
\hline
\multicolumn{1}{|c|}{} & Monolingual Arabic & Monolingual English & Arabic-English CS & Total     \\ \hline
Facebook               & 634,914            & 140,954             & 61,210            & 837,078   \\ \hline
Twitter                & 886,904            & 129,787             & 62,235            & 1,078,926  \\ \hline
Facebook + Twitter                  & 1,521,818          & 270,741             & 123,445           & 1,916,004 \\ \hline
\end{tabular}
\caption{The number of sentences gathered from Facebook and Twitter per language.}
\label{table:sm_corpus}
\end{table}
% Placed here to have the figure on top on page 2. Couldn't fit it any other way.
\section{Bilingual Word Embeddings}
\label{sec:approach}

We train bilingual word embeddings using three of the models compared in~\cite{UFD+16} and propose two other simple extensions (BiCCAonBiSkip and Bi-CS). In order to conduct a fair comparison between all algorithms, the same data is used for training all word embeddings. Across all models, both corpora are used: the set of Egyptian Arabic parallel sentences in~\cite{ZMD+12} as well as the text corpus that was gathered from the social media platforms.

 \begin{figure*}[h]
  \centering
    \includegraphics[width=1.0\textwidth]{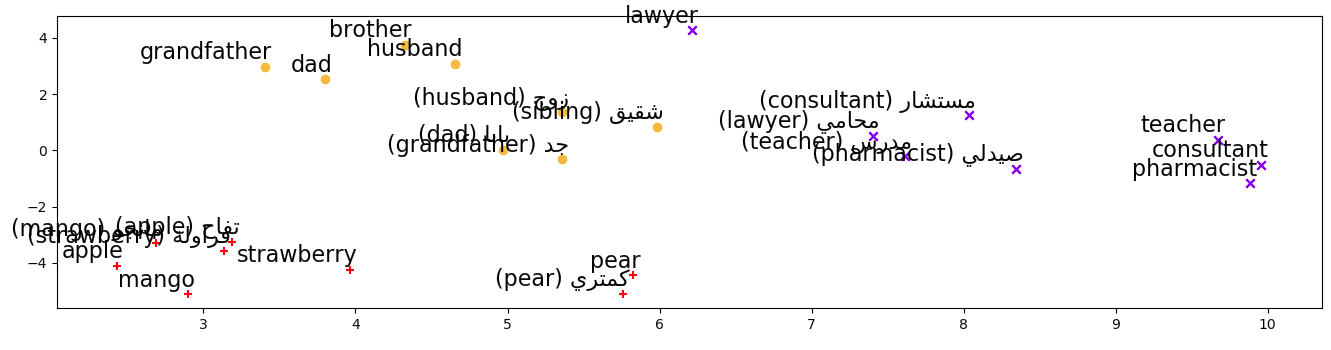} \hfil
    \includegraphics[width=1.0\textwidth]{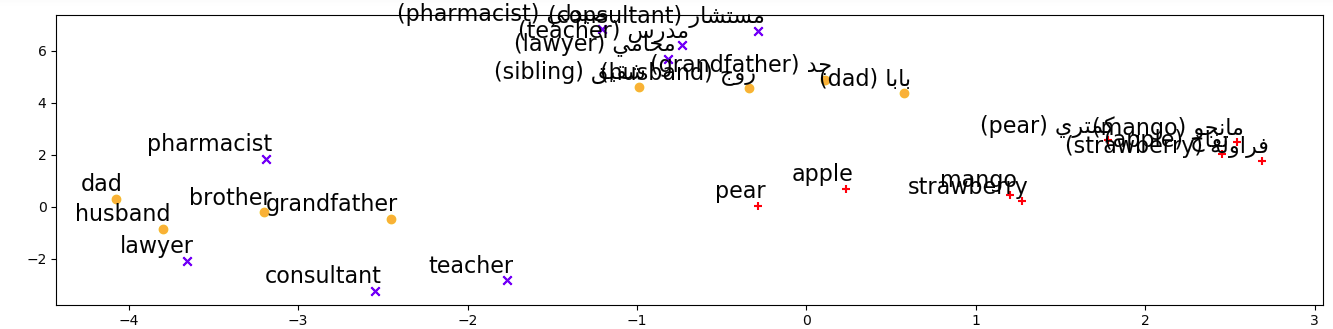}
      \caption{t-SNE~\cite{MH08} visualization of BiCCAonBiSkip and Bi-CS for words in 3 clusters: fruits (+), family (o) and professions (x).}
    \label{fig:biCCA_biSkip}
\end{figure*}

\paragraph{Bilingual Compositional Vector Model (BiCVM)}
Hermann and Blunsom~\cite{HB14} proposed to use sentence-aligned parallel data to train bilingual word embeddings. Their model is motivated by the fact that aligned sentences express the same meaning and therefore have similar sentence representation. We train the models using the parallel corpus, as well as the text corpus, where each sentence in the text corpus acts as its own equivalent sentence. The models are trained using both proposed composition functions used for summarizing a sentence; additive (BiCVM\_add) and bigram (BiCVM\_bi) models. We train the models using a hinge loss margin set to the embeddings dimension ($m=d$, as set in~\cite{HB14}), noise parameter of 10, batch size of 50,  L2 regularization with $\lambda=1$, a step-size of 0.05 and \textit{AdaGrad} as the adaptive gradient method~\cite{DHS11}. All models are trained over 20 iterations.

\paragraph{Bilingual Correlation-Based Embeddings (BiCCA)}
Faruqui and Dyer~\cite{FD14} proposed to project two (independently trained) vector spaces into a single space, with the use of a translation lexicon. We investigate the use of continuous bag-of-words (CBOW) and Skip-gram models~\cite{MCC13} for the vector spaces before projection. The Arabic vector space is trained using the subset of Arabic and CS sentences in the text corpus as well as the parallel corpus, while the English vector space is trained using the subset of English and CS sentences in both corpora. CBOW models are trained with Negative Sampling, while Skip-gram models are trained with Hierarchical Softmax. All models are trained with a window of size 5. For BiCCA projection, we set the number of canonical components $k=0.8$.

\paragraph{Bilingual Skip-gram Model (BiSkip)}
Luong et al.~\cite{LPM15} proposed BiSkip to train bilingual embeddings using parallel corpora with word alignments. Given a word alignment link from $w_1$ in language $l_1$ to $w_2$ in language $l_2$, the model predicts the context words of $w_1$ using $w_2$ and vice-versa. The model aims at creating high-quality monolingual and bilingual embeddings by learning the context concurrence information from the monolingual text, and the meaning equivalent signals from the parallel corpus. As in BiCVM, the parallel and text corpora provide the sentence-aligned data used for training the word embedding models. We use cdec \cite{DWS+10} to lowercase, tokenize and generate word alignments for the parallel corpus. For the sentences obtained from the text corpus, we use fake alignments, where each word is aligned with itself. The models are trained with the same hyperparameters as given in \cite{UFD+16}. We use a window of size 10, cross-lingual weight of 4, 30 negative samples, and 5 training iterations.

\paragraph{BiCCAonBiSkip}
We observed that BiSkip model outperforms BiCCA and BiCVM in monolingual and bilingual intrinsic evaluations. However, BiCCA provides a vector space where similar words from both languages were most closely represented. Therefore, we combine both best-performing approaches by applying BiCCA on the output embeddings of BiSkip to draw the Arabic and English word embeddings closer, while maintaining the high monolingual and bilingual quality of BiSkip embeddings.

\paragraph{Bilingual CS Embeddings (Bi-CS)}
It is not always possible to find parallel corpora or bilingual lexicons, especially for low-resource languages.
In this work, CS data is available for free which could be used as glue forcing the monolingual embeddings being closer together in the shared vector space.
Therefore, we propose Bi-CS that jointly learns word representations of both languages by training the Skip-gram and CBOW models on monolingual data of both languages in combination with a small amount of CS data. 
The models are trained using the concatenation of all sentences in the text and parallel corpora. The parallel corpus was used as a text corpus added to the training data in order to unify the data used across algorithms to insure a fair comparison. However, the Bi-CS approach does not require any parallel data. CBOW models are trained with Negative Sampling, while Skip-gram models are trained with Hierarchical Softmax. All models are trained with a window of size 5.

\section{Language Modeling}
\label{sec:LM}

\paragraph{Neural-based LM} We use TheanoLM\footnote{\url{https://github.com/senarvi/theanolm}} (TLM) to train our recurrent neural network (RNN) LMs~\cite{EK16}. As TLM does not allow the use of pretrained word embeddings, we extend the implementation accordingly by adding an embedding layer that is initialized with pre-trained embeddings and replaces the projection layer in the small default architecture in~\cite{EK16}. The embedding layer weights are then updated with the other layers. A neural language model with an LSTM layer, followed by a softmax output layer is used. The embedding layer is of size 100 or 200, the LSTM layer has 300 LSTM cells and the softmax layer has the vocabulary size. We use this network architecture in all experiments.  All the bilingual embeddings are filtered to only include words from the training data to ensure an identical vocabulary across all experiments and comparable PPLs. For all experiments, we fix the random seed, and optimize with \textit{AdaGrad}~\cite{DHS11}. 
We set the mini-batch size to 16 and train until no improvement is observed on the development set.
\paragraph{Experiment 1} In the first setting, the models are trained using the extended speech transcriptions \cite{HEA18}. The corpus was divided into training, development and testing sets as follows: 2040, 187 and 180 sentences. The division was made taking into consideration having balanced sets in terms of speakers and genders. The bilingual embeddings are filtered to only include words from the training set.
\paragraph{Results} In Table \ref{table:ppl}, we compare the PPLs obtained using the bilingual word embeddings for the first experiment. We only show the results obtained using embedding dimension of 200, as they are superior over the dimension of 100. The baseline LM with randomly initialized embeddings has a PPL of 300.2 on the development set and 291.5 on the evaluation set. All bilingual word embeddings outperform the baseline with a large margin. Despite the simplicity of the Bi-CS approach, it achieves best results, giving a relative improvement of 33.5\% in PPL on the test set.
\paragraph{Experiment 2} 
In experiment 1, the substantial improvements in PPLs show great potential in the integration of bilingual word embedding for improving CS speech recognition. However, under this setting, the improvements in the PPLs cannot be traced back exclusively to the incorporation of the pre-trained word embeddings, as the LMs using the pre-trained word embeddings were given access to more CS data over the baseline. 
Therefore, we conduct a second experiment where all models are trained using the CS sentences obtained in our social media text corpus. The word embeddings are also filtered to only contain the words in that subset of the corpus. The Egyptian Arabic speech transcriptions obtained in~\cite{HEA18} are used in development and testing, where the corpus was divided evenly into two sets.
\paragraph{Results} In Table \ref{table:ppl}, we compare the PPLs obtained using the bilingual word embeddings for the second experiment. The results are similar to those of the first experiment, where all bilingual word embeddings outperform the baseline, and Bi-CS gives the lowest PPL. The best-performing (Bi-CS) model achieves a relative improvement of 22.4\% in PPL on the test set. It is to be noted that in this setting, the PPLs are very high since the models are trained on text gathered from social media platforms, while it is tested on speech text. This setting is only used to confirm the effectiveness of using bilingual word embeddings in low-resourced CS language modeling.

\begin{table} \centering
\begin{tabular}{l|cc|cc}
            &\multicolumn{2}{c|}{\textbf{Experiment 1}} &\multicolumn{2}{c}{\textbf{Experiment 2}}\\\hline
            &\textbf{Dev}&\textbf{Test}&\textbf{Dev}&\textbf{Test} \\
            \hline
Baseline        & 300.2             & 291.5         & 2146.7            &2188.0 \\
\hline
BiCVM\_add      &226.3              &228.1          &1890.9             &1994.1  \\
BiCVM\_bi       &225.3              &217.7          &1977.4             &2102.9  \\
BiSkip          &224.9              &220.9          &1712.1             &1851.5   \\
BiCCA\_skip     &249.5              &241.4          &2033.6             &2156.6   \\
BiCCA\_cbow     &247.8              &241.4          &1956.8             &2145.2   \\
\hline
BiCCAonBiSkip   &257.5              &248.9          &1988.9             &2117.8   \\
Bi-CS\_skip     &223.2              &214.5          &1895.2             &2016.6  \\
Bi-CS\_cbow     &\textbf{204.7}     &\textbf{193.6} &\textbf{1588.3}    &\textbf{1697.0}   \\
\hline
\end{tabular}
\caption{Experiment 1 - PPLs on the development and the test set}
\label{table:ppl}
\end{table}

\section{Intrinsic Evaluations}
\label{sec:eval}

In this Section, we present intrinsic evaluations on the word embeddings. The goal of bilingual word embeddings is to obtain distributed word representations that are of high quality monolingually and bilingually; such that similar words in each language and across different languages end up close to each other in the embedding space. We evaluate both intrinsic objectives: (a) monolingual objective using \textit{concept categorization} task on Arabic and English words separately and (b) bilingual objective using 
\textit{concept categorization} on words from both languages. 

\paragraph{Concept Categorization} The task of \textit{concept categorization}, also known as \textit{word clustering}, is to divide a set of words into $n$ subsets. (e.g., correctly categorizing the words in the set \{\textit{teacher},\textit{apple},\textit{mango},\textit{scientist}\} into two subsets)
The word vectors are clustered into $n$ groups (where $n$ is determined by the gold standard partition) using the CLUTO toolkit~\cite{Kar03}. Clustering is done with the \textit{repeated bisections} with \textit{global optimization} method and CLUTO's default settings otherwise, as outlined in~\cite{BDK14}. Performance is evaluated in terms of \textit{purity}, which measures the extent to which each cluster contains words from primarily one category, as defined by the gold standard partition. We created our own gold standard partition using 12 concept categories from the compiled thesaurus. The dataset consists of 211 Egyptian Arabic and 205 English words (or ``concepts''). In order to evaluate the word embeddings in terms of monolingual quality, we report the clustering \textit{purity} of the Arabic words and English words separately. The overall monolingual \textit{purity} is calculated as the average of both. For the bilingual evaluation, we report the \textit{purity} of clusters on all words from both languages. The results are shown in Table \ref{table:intrinsicEval}.\

\begin{table} \centering \small
\begin{tabular}{lcccc}
            &\textbf{ARB}&\textbf{EN}&\textbf{Mono.}&\textbf{Bi.} \\
            \hline
BiCVM\_add          &32.2\%         &24.9\%             &28.6\%             &27.2\% \\
BiCVM\_bi           &26.5\%         &23.9\%             &25.2\%             &23.1\% \\
BiSkip              &64.5\%         &38.0\%             &51.3\%             &41.6\% \\
BiCCA\_skip    &42.2\%         &\textbf{57.6\%}    &49.9\%             &\textbf{43.0\%} \\
BiCCA\_cbow         &36.0\%         &52.7\%             &44.4\%             &38.2\% \\

\hline
BiCCAonBiSkip       &\textbf{70.6\%}&46.8\%             &\textbf{58.7}\%    &\textbf{43.0}\% \\
Bi-CS\_skip    &56.4\%         &48.8\%             &52.6\%             &42.3\% \\
Bi-CS\_cbow         &57.8\%         &46.8\%             &52.3\%             &\textbf{43.0}\%  \\
\hline
\end{tabular}
\caption{Intrinsic evaluations of bilingual word embeddings (Mono. = monolingual and Bi. = bilingual)}
\label{table:intrinsicEval}
\end{table}

\paragraph{Results} The performance of the three models (BiCVM, BiSkip and BiCCA) shows that BiSkip and BiCCA outperform BiCVM on all tasks. The ordering of BiSkip and BiCCA is task-dependent: BiSkip achieves better results on Arabic monolingual task, while BiCCA performs better on the English monolingual task as well as the bilingual task. By combining both approaches, the BiCCAonBiSkip model outperforms all other models for both monolingual and cross-lingual categorization tasks. This shows that BiCCAonBiSkip is able to bring the crosslingual word vectors closer without compromising the high-quality embeddings provided by BiSkip.
Figure \ref{fig:biCCA_biSkip} presents a t-SNE~\cite{MH08} visualization of word embeddings obtained by BiCCAonBiSkip for words in three clusters: fruits, family and professions, in which words from both languages can be observed within clusters.
Surprisingly, Bi-CS, which is trained in a completely unsupervised manner, displays good performance, outperforming BiSkip, BiCCA and BiCVM  which rely on richer resources. This is an interesting observation, as it is usually the case that models trained with weaker supervision show lower performance on semantic tasks \cite{UFD+16}. 

\section{Discussion}
\label{sec:discussion}
While all models show varying performance across tasks, Bi-CS achieves good performance consistently, outperforming all models in LM, and performing second best on the categorization task. This highlights the effectiveness of the Bi-CS model, especially that it requires the least data and supervision requirements. When comparing the performances of the other models, we find that the their ordering is inconsistent within categorization, as well as across categorization and LM tasks. This observation is in-line prior work \cite{SLM+15,UFD+16,GLR+19} reporting that performances are task-dependent. It is interesting to note that the models show almost opposite ordering of performances (excluding Bi-CS). The performance mismatch is most obvious for BiCCAonBiSkip (achieving highest performance in categorization and least in LM) and BiCVM (ranking last in categorization and second best in LM). The mismatch can also be seen in the additive and bigram composition functions used by BiCVM, where the former performs better in the categorization tasks, while the latter shows better results in LM. Furthermore, when comparing CBOW and Skip-gram models, we find that Skip-gram models (mostly) outperform CBOW models in the categorization task, while CBOW models are superior in LM. It is also to be noted that word embeddings fuse multiple word senses (or meanings) into one representation. Given that sense embeddings have shown improvements in  NLP tasks \cite{LJ15,CAC17}, it would be interesting to further improve our embeddings to incorporate the different word senses and investigate its effect on the categorization and CS LM tasks.

\section{Conclusion}
\label{sec:conc}
We investigated the use of various state-of-the-art bilingual word embeddings for improving CS LM. 
We explored various state-of-the-art approaches that require different levels of cross-lingual supervision for training the embeddings. 
In order to relax the need for parallel corpora and bilingual lexicons, which are usually scarce, we proposed Bi-CS, a simple, yet effective model. Bi-CS only requires monolingual corpora along with a small amount of CS data, and can thus be more easily applied to low-resource languages. All LMs using bilingual word embeddings outperformed the baseline trained with randomly initialized word embeddings. Bi-CS gives the best performance, achieving a relative improvement of 33.5\% over the baseline. It also outperforms the existing approaches on the intrinsic evaluation. In future work, we plan to investigate the effectiveness of incorporating the bilingual word embeddings into language modeling on the task of automatic speech recognition.

\bibliographystyle{splncs04} 
\bibliography{references}

\end{document}